\documentclass[10pt,twocolumn,letterpaper]{article}

\usepackage{3dv}
\usepackage{times}
\usepackage{epsfig}
\usepackage{graphicx}
\usepackage{amsmath}
\usepackage{amssymb}
\usepackage{algorithm}
\usepackage{algorithmic}

\usepackage[pagebackref=true,breaklinks=true,letterpaper=true,colorlinks,bookmarks=false]{hyperref}

\threedvfinalcopy 

\def\threedvPaperID{} 

\ifthreedvfinal\fi
\begin{document}

\title{Cross-modal Center Loss}

\author{Longlong Jing \quad Elahe Vahdani \quad Jiaxing Tan \quad Yingli Tian\\
City Univerisity of New York\\
}


\maketitle

\begin{abstract}

Cross-modal retrieval aims to learn discriminative and modal-invariant features for data from different modalities. Unlike the existing methods which usually learn from the features extracted by offline networks, in this paper, we propose an approach to jointly train the components of cross-modal retrieval framework with metadata, and enable the network to find optimal features. The proposed end-to-end framework is updated with three loss functions: 1) a novel cross-modal center loss to eliminate cross-modal discrepancy, 2) cross-entropy loss to maximize inter-class variations, and 3) mean-square-error loss to reduce modality variations. In particular, our proposed cross-modal center loss minimizes the distances of features from objects belonging to the same class across all modalities. Extensive experiments have been conducted on the retrieval tasks across multi-modalities, including 2D image, 3D point cloud, and mesh data. The proposed framework significantly outperforms the state-of-the-art methods on the ModelNet40 dataset. 

\end{abstract}

\section{Introduction}

\begin{figure}[tb]
	\centering
	\includegraphics[width = 0.5\textwidth]{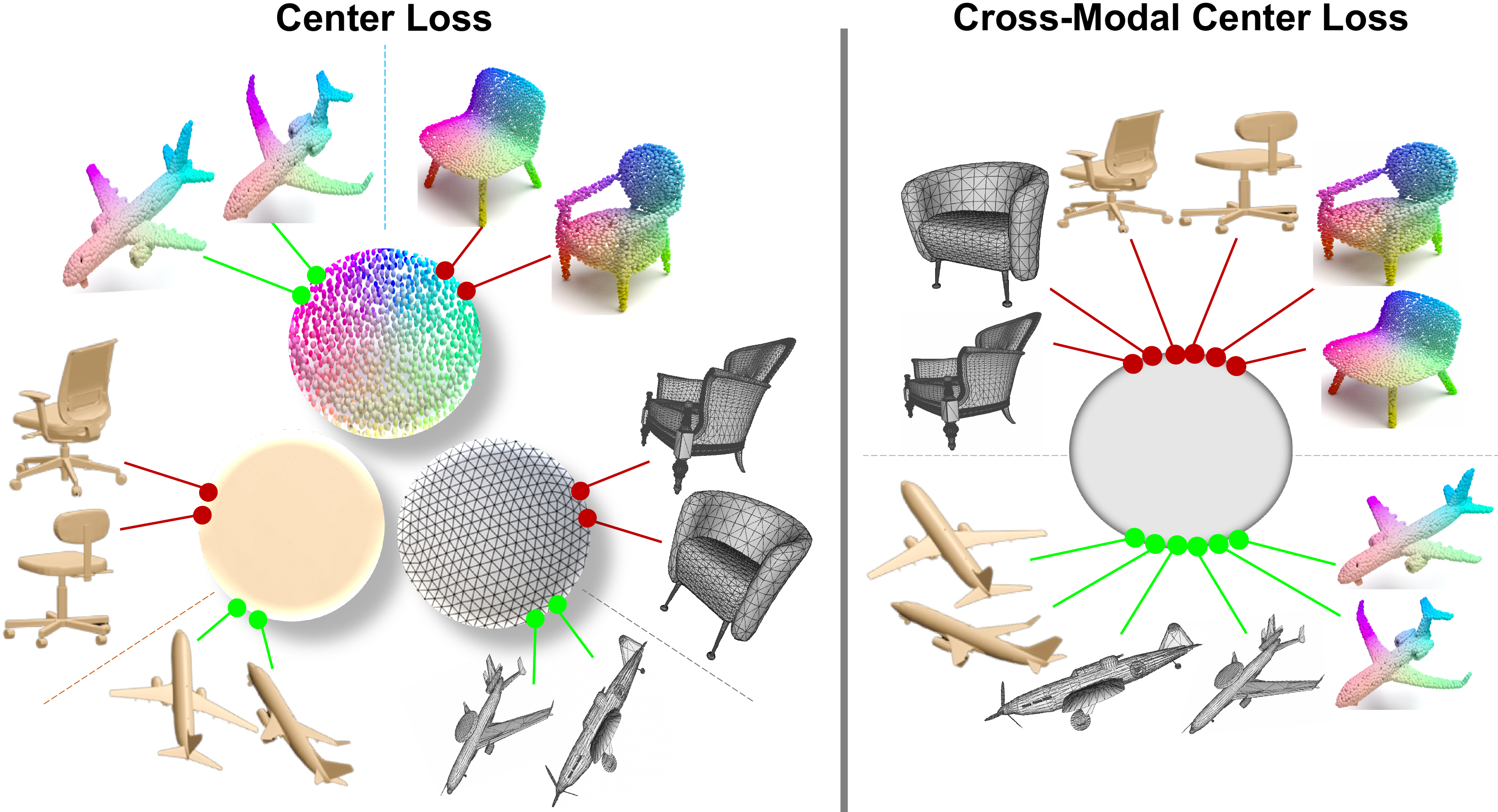}
	\caption{
	Traditional center loss vs. the proposed cross-modal center loss. Our proposed cross-modal center loss (right) finds a unique center for each class across all modalities. Traditional center loss (left) finds a center for each modality and each class and ignores the relation among centers of different modalities. Our proposed cross-modal center loss specifically eliminates the discrepancy across multiple modalities and thus is very effective for learning modal-invariant features.}
	\label{fig:motivation}
\end{figure}

With the stream of multimedia data flourishing on the Internet in different formats such as videos, images, or text, cross-modal retrieval task has attracted more and more attention from the multimedia communities. Cross-modal retrieval is the task of retrieving data from one modality given a query from a different modality. Inspired by the representation power of deep learning, a series of deep learning-based methods have been proposed for cross-modal retrieval \cite{nagrani2018seeing, zhou2018visual, zhen2019deep}. These methods operate by learning modal-invariant representations in a common space.

The features from different modalities generally have different distributions. Therefore, a fundamental requirement for the cross-modal retrieval task is to bridge the gap among different modalities, commonly done by representation learning. The existing methods mainly extract each modality's features by offline pre-trained models and apply a projection function to transfer the features into a common representation space. By this transformation, the similarity of features from different modalities can be directly measured. Hence, the main challenge during this process is to learn discriminative and modal-invariant features.

By learning discriminative features, we ensure that data from the same class are mapped closely to each other in the feature space while different classes are separated as far as possible. In many studies, cross entropy or mean square error loss in the label space is used to maximize the inter-class variations. In order to compare the features extracted from different modalities, the features need to be modal-invariant. Various methods are proposed to reduce the cross-domain discrepancy by using adversarial loss, sharing a projection network, using triplet loss with pairs/triplets of different modalities, maximizing cross-modal pairwise item correlation. To preserve cross-modal semantic structures, Peng \textit{et al.} \cite{peng2017ccl} proposed to optimize both intra-modality and inter-modality correlation feature representations in a common space. Wang \textit{et al.} \cite{wang2017adversarial} pointed out that pairwise correlation is not sufficient and proposed to adapt the triplet loss \cite{schroff2015facenet} to minimize the intra-class distance while maximizing inter-class distances for all samples from different modalities based on their labels. Khosla \textit{et al.} \cite{khosla2020supervised} proposed a supervised version of contrastive loss \cite{hadsell2006dimensionality} to push apart clusters of samples from different classes while samples of the same class are mapped closely to each other. However, when dataset size is large, the number of pairs/triplets created in training will drastically increase, which results in slow convergence and instability in the training \cite{wen2016discriminative}.

Although these methods achieved promising results in the cross-modal retrieval tasks, they suffer from the following limitations: 1) Their core idea is to minimize the cross-modal discrepancy over the features from multiple modalities extracted by pre-trained neural networks. For example, in the task of image-text retrieval, image and text features are extracted by ImageNet pre-trained models (e.g., VGG) and SentenceCNN, and then learning is performed on these extracted features instead of the metadata. Because these feature extractors (VGG, SentenceCNN) are not trained or finetuned for the cross-modal retrieval task, they are not optimally representative. Instead, the network should be jointly trained with multimodal data to address the retrieval task appropriately. 2) The existing loss functions are mainly designed for two types of modalities, image, and text, and may not generalize well for cases when more than two modalities are available. It is essential to develop a simple yet effective loss function that can be easily extended for multiple modalities.

In this paper, we propose a new loss function, called Cross-modal Center Loss, specifically designed to minimize the intra-class variation across multiple modalities. Our loss function is directly inspired by the traditional uni-modal center loss, which learns a center for each class and minimizes the distance between objects and their corresponding centers in the feature space. Fig.~\ref{fig:motivation} shows the comparison between the traditional center loss and the newly proposed cross-modal center loss. The traditional center loss minimizes the distance of objects and their centers in separate features spaces defined for each modality. Instead, our proposed cross-modal center loss learns a unique center \textit{C} for each class in the common space of all modalities. Explicitly, it minimizes the distance of multi-modal objects and their centers in the same common feature space for all modalities. When more multi-modal data is available, the cross-modal center loss will be able to learn more reliable centers for each class in the common space.

With the proposed cross-modal center loss, the cross-modal discrepancy between different modalities of the data can be eliminated. The proposed cross-modal center loss can be employed in conjunction with other loss functions to learn the features for cross-modal retrieval jointly. To verify the effectiveness of the proposed loss function, we further propose an end-to-end framework for cross-modal retrieval task to learn discriminative and modal-invariant features. The proposed framework is optimized with three loss functions, including the cross-entropy in the label space to learn discriminative features, the cross-modal center loss to eliminate the cross-modal discrepancy in the universal space, and the mean square error loss to minimize the cross-modal distance per object. Furthermore, a weight sharing strategy is applied to learn modal invariant features in the common space.

Unlike previous cross-modal retrieval methods that extract the features of image or text by offline networks, we propose to jointly train the entire framework from the metadata without being limited by pre-trained models from other datasets. The effectiveness of the proposed framework is evaluated on a novel 3D cross-modal retrieval task, which has not been explored by existing supervised methods. Our method significantly outperforms the recent state-of-the-art methods on the 3D cross-modal retrieval task. The main contributions of this paper are summarized as follows:
\begin{itemize}
\item We propose a novel cross-modal center loss to map the representations of different settings of modalities into a common feature space.
\item We propose an end-to-end framework for the cross-modal retrieval task by jointly training multiple modalities using the proposed cross-modal center loss. The proposed framework can be extended to various cross-modal retrieval tasks.
\item The proposed framework significantly outperforms the state-of-the-art methods on cross-modal retrieval across images, point cloud, and mesh for 3D shapes. To the best of our knowledge, this is the first supervised learning method for object retrieval across the 2D image, 3D point cloud, and mesh data.
\end{itemize}

\section{Related Work}

\noindent \textbf{Feature Learning for 3D Objects}: 3D data are inherently multi-modal and can be represented in various ways such as point cloud, multi-view images, mesh, volumetric data, and other forms. Various deep learning-based methods have been proposed for 3D feature learning including unordered point cloud-based methods \cite{lei2019spherical,li2019deepgcns,qi2017pointnet,qi2017pointnet++,thomas2019kpconv,wang2019dynamic,wu2019pointconv,XMV}, multi-view images-based methods \cite{su2015multi,su2018deeper,XMV}, and volumetric voxelized data-based methods \cite{chang2015shapenet,volum4,volum5,volum1,volum3}. Qi \textit{et al.} proposed the first deep learning-based model (i.e., PointNet) to directly learn the features from unordered point cloud data. To specifically model the local information for each point \cite{qi2017pointnet}, Wang \textit{et al.} proposed a dynamic graph convolution neural network (DGCNN) with EdgeConv using $k$ nearest neighbor (KNN) points \cite{wang2019dynamic}. Su \textit{et al.} proposed to learn the features for 3D objects with multi-view CNN operating on 2D images that are rendered from different views of 3D data \cite{su2015multi}. MeshNet \cite{feng2018meshnet} and MeshCNN \cite{hanocka2019meshcnn} were proposed to learn features directly from the mesh data by modelling the geometric relations of mesh faces of the object. Recently, few studies attempted to learn modal-invariant features with self-supervised learning \cite{SelfSurvey}. Jing \textit{et al.} proposed MVI for unsupervised modal and view-invariant feature learning by contrasting where the learned features can be used for cross-modal retrieval \cite{MVI}.

\noindent \textbf{Cross-modal retrieval:} Several methods have been proposed for the cross-modal retrieval task, mainly targeting image-text retrieval. One straightforward solution for this task is to formulate the problem as a linear projection. Canonical Correlation Analysis (CCA) is one of the most typical approaches for this problem, which learns the linear projection by maximizing the correlation between two modality sets \cite{hotelling1992relations}. As an extension from bi-modality to the multi-modality relationship, Kan \textit{et al.} proposed multi-view discriminant analysis (MvDA) to jointly learn multiple view-specific linear transformations as a generalized Rayleigh quotient optimization problem \cite{kan2015multi}. Moreover, a scalable multi-view canonical correlation analysis (MCCA) was proposed by Rupnik \textit{et al.}, which reduces the time complexity from cubic to linear \cite{rupnik2010multi}. In addition, a non-linear kernel-based extension was also discussed in this work. To utilize semantic information, Zhai \textit{et al.} proposed a noise-resistant joint representation learning (JRL) model to jointly learn the correlation and semantic information in a semi-supervised manner \cite{zhai2013learning}.

Most recently, deep learning-based methods have been proposed for representation learning due to their feature learning power. Inspired by CCA, Andrew \textit{et al.} proposed deep canonical correlation analysis (DCCA) to adapt deep neural networks to model the complex nonlinear transformations by projecting two highly linear correlated views into the same common space \cite{andrew2013deep}. Furthermore, Wang \textit{et al.} proposed deep canonically correlated autoencoders (DCCAE). This two-autoencoder design is jointly optimized by combining the canonical correlation between the learned representations and the reconstruction errors of the autoencoders \cite{wang2015deep}. Peng \textit{et al.} proposed a two-stage framework called Cross-Media Deep Networks (CMDM), which acquires inter- and intra-modality features and then hierarchically combines the representations to further learn the rich cross-media correlations \cite{peng2016cross}. However, these deep learning-based methods did not concentrate on inter- and intra-modality relations in their designs. The CMDN later was extended by Peng \textit{et al.} to cross-modal correlation learning (CCL) by adding inter-modal interactions in the first stage while adding intra-modal semantic constraints in the second stage \cite{peng2017ccl}. 

To learn modal-invariant features, Wang \textit{et al.} proposed adversarial cross-modal retrieval (ACMR) which adapted adversarial learning to minimize the domain gap by using a discriminator to predict the corresponding modality of the representations \cite{wang2017adversarial}. With the adversarial loss function, this method significantly outperformed the previous state-of-the-art methods on popular benchmarks with a large margin. Zhen \textit{et al.} proposed deep supervised cross-modal retrieval (DSCMR) to learn the representations in the common space in regard to both inter-class and intra-class relations \cite{zhen2019deep}. DSCMR increases the inter-class variations via the discrimination loss in both the label space and the common representation space. Moreover, DSCMR reduces the cross-modal discrepancy by minimizing the modal-invariant loss in the feature space.

Most of the existing work use the image and text features extracted by offline networks and directly minimize the cross-modal gap in the common space using these features. In this paper, we propose an end-to-end jointly trained framework and a novel cross-modal center loss to learn discriminative and modal-invariant features directly from metadata. 

\section{Methods}

We propose an end-to-end framework with joint training of multiple modalities for cross-modal retrieval task based on the proposed cross-modal center loss. The formulation of the proposed cross-modal center loss and its application for cross-modal retrieval tasks are introduced in the following sections.

\subsection{Problem Formulation}


Dataset $S$ contains $N$ instances where the $i$-th instance $t_i$ is a set of $M$ modalities with a semantic label $y_i$. The set of modalities of $t_i$ is denoted by $s_i$. Formally:

\begin{center}
$S= \{t_i\}_{i=1}^{N}$ , \hspace{0.1in} $t_i = \big (s_i, y_i \big ) $ , \hspace{0.1in} $s_i= \{ x_{i}^{m}\}_{m=1}^{M}$ 
\end{center} 

Generally, the modality samples $\{x_{i}^{1},x_{i}^{2}, \cdots, x_{i}^{M} \}$ are in $M$ different representation spaces and their similarities cannot be directly measured. The goal of the cross-modal retrieval task is to learn $M$ projection functions $f_m$ for each modality $m \in[1, M]$, where $v^{m}_{i}=f_{m}(x_{i}^{m},\theta_{m})$ and $\theta_{m}$ is a learnable parameter. As a result, $v^{m}_{i}$ is a projected feature in the common representation space. Distance between the projected features is a measure of similarity between the samples across all modalities. Therefore, samples from the same class should be mapped closely to each other independent of their modalities: $d(v^{m}_{i} , v^{m^{\ast}}_{i}) \sim \text{low}$. On the other hand, samples from different classes should be projected as far as possible:  $d(v^{m}_{i} , v^{m^{\ast}}_{j}) \sim \text{high}$ (where $i \neq j$).

\begin{figure*}[tb]
\centering
\includegraphics[width = 0.7\textwidth]{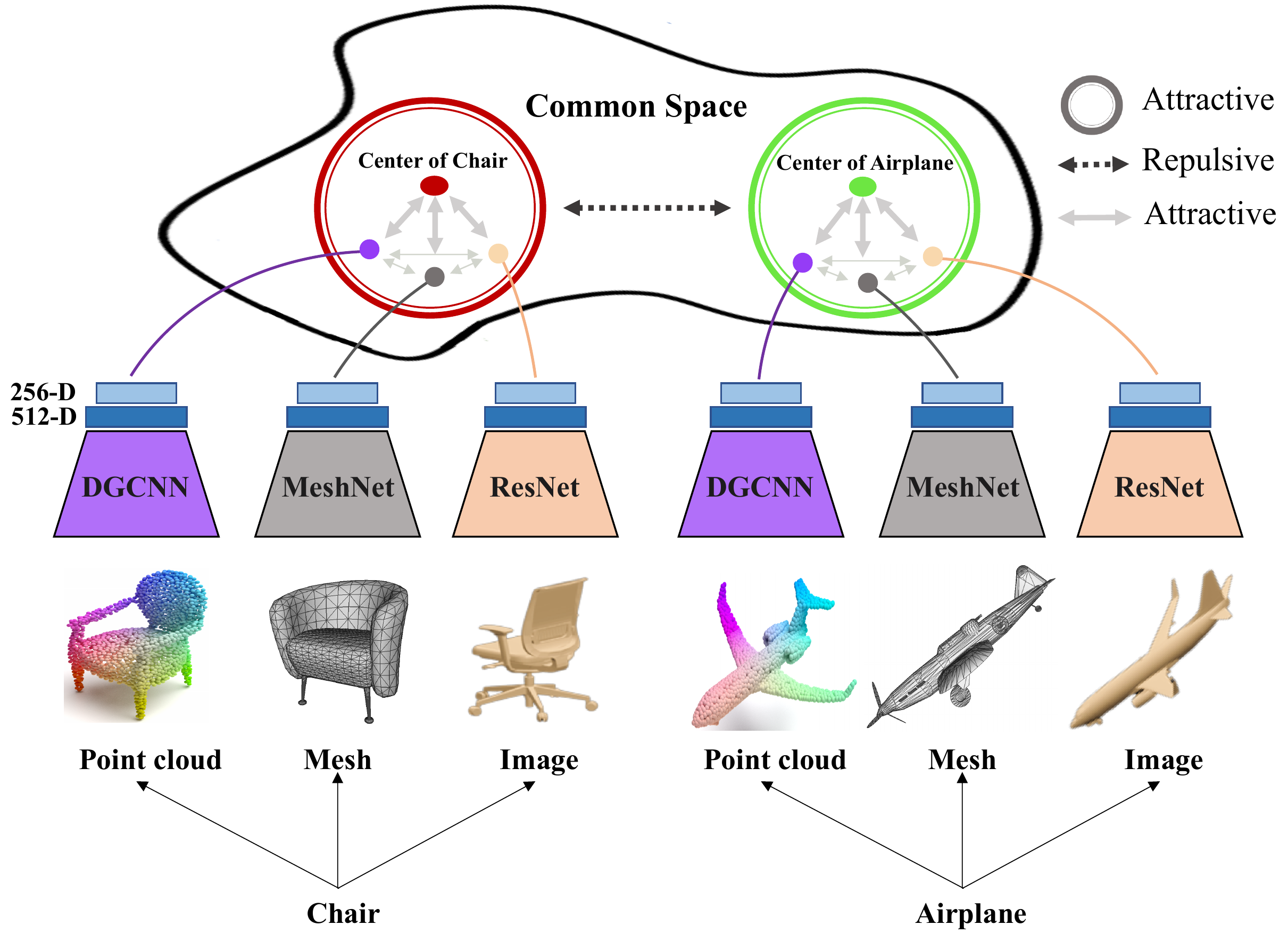}
\caption{An overview of the proposed framework for 3D cross-modal retrieval task. Mesh, point cloud, and multi-view 2D image features are extracted by MeshNet, DGCNN, and ResNet, respectively, then projected to a common space via two shared fully connected layers. With the cross-modal center loss in conjunction with the cross-entropy loss and mean square error loss, the proposed framework can learn discriminative and modal invariant features.}
\label{fig:framework}
\end{figure*}

\subsection{Loss Function}

The core of the cross-modal retrieval is to obtain discriminative and modal-invariant features for data of different modalities with heterogeneous networks. To learn discriminative features, we use the cross entropy loss over the sharing head of our network, while our proposed cross-modal center loss and mean square error help with learning modal-invariant features.

\textbf{Cross-modal center Loss:} Given the extracted features $ \{ v^{m}_i \} _{i = 1}^{N}$ ($m\in [1, M]$) for $N$ instances and $M$ modalities, our proposed cross-modal center loss is formulated in Eq. \ref{eq:center-def}: 

\begin{equation}
\label{eq:center-def}
L_{c}=\frac{1}{2}\sum_{i=1}^{N}\sum_{m=1}^{M}\left \| v^{m}_i - C_{y_i} \right \|_2^2,
\end{equation}
\noindent

where $C_{y_i} \in \mathbb{R}^k$ denotes the center of class $y_i$ in the common space and $k$ is the dimension of features. Comparing to the original center loss \cite{wen2016discriminative}, our proposed cross-modal center loss learns by eliminating the cross-modality gap and reducing the intra-class variation. To learn modal-invariant features, the cross-modal center loss optimizes the network to learn a center $C_{y_i}$ for class $y_i$ and minimize the distance between the features and their corresponding centers within each training batch. After each training iteration, the center of each class, $C_{j}$, is updated by $\bigtriangleup C_j$ with data from all modalities belonging to class $j$:

\begin{equation}
\label{eq:center}
\bigtriangleup C_j=\frac{\sum_{i=1}^{N}\sum_{m=1}^{M}\delta(y_i=j)(C_j-v^{m}_i)}{1+\sum_{i=1}^{N}\delta(y_i=j)},
\end{equation}
\noindent   
where

\begin{equation}
    \delta(condition)=\left\{\begin{matrix}
1 & condition = True \\ 
 0& otherwise 
\end{matrix}\right.
\end{equation}

Given a large batch size, the model can learn a robust center for each class, leading to produce features with small intra-class variation across all modalities. One advantage of the proposed cross-modal center loss is that it can be easily extended to more modalities. When data with more modalities are available, it provides more robust centers and may lead to better optimized features.

\textbf{Discriminative Loss:} To learn discriminative features, cross entropy loss in the label space is employed to optimize the network. Given $N$ samples from $M$ modalities, the discriminative loss is calculated by the cross-entropy loss between the predicted probability $\hat{y^{m}_i}$ and its label $y_i$.

\begin{equation}
    L_{d}=-\frac{1}{N}(\sum_{i=1}^{N}\sum_{m=1}^{M}y^{m}_i\cdot log(\hat{y^{m}_i}) ), 
\end{equation}
\noindent  
where $\hat{y^{m}_i}$ is predicted by two shared layers applied over the extracted feature $v^{m}_i$: 


\begin{equation}
      \hat{y^{m}_i} = MLP(v^{m}_i).
\end{equation}

Trained with cross-entropy loss, samples from the same category have higher similarities, while samples from different categories have lower similarities. Jointly trained with cross-modal center loss and cross-entropy loss, the network is able to learn both modal-invariant and discriminative features.

To further reduce the cross-modal discrepancy for each instance, we propose a loss function based on mean square error to minimize the distances between the features of all cross-modal sample pairs. The loss function across $M$ modalities for each instance $i$ is defined as the following where $\{ v^{1}_i, v^{2}_i, \cdots, v^{M}_i \}$ are the extracted features: 

\begin{equation}
   L_m=   \sum _{ \alpha, \beta \in  [1,M] \alpha \neq \beta } \left \| v^{\alpha}_i - v^{\beta}_i \right \|_2^2.
\end{equation}

The three proposed loss functions are used to jointly train the network to learn discriminative and modal-invariant features:

\begin{align}\label{eq:loss_combined}
Loss = \alpha_c L_{c} +\alpha_d L_{d} + \alpha_m L_{m},
\end{align}
\noindent
where $\alpha_c$, $\alpha_d$, and $\alpha_m$ are the hyper-parameters designed for each loss term. Our proposed joint loss function in Eq. \ref{eq:loss_combined} can be optimised by stochastic gradient descent. The details of the optimization procedure is summarized in \textit{Algorithm \ref{algo:optimize}}.

\begin{algorithm}
 \caption{Optimization procedure of the proposed framework}
 \begin{algorithmic}[1]
 \label{algo:optimize}
 \REQUIRE The training data set $S=\left \{(t_i, y_i)\right \}_{i=1}^{n}$, the dimensionality of the common representation space $k$, the mini-batch size $n_b$, the learning rate $\tau$ , the maximal number of epochs $\mathcal{N}$.
 \ENSURE The optimized parameters in $M$ sub-networks $\theta_{m}, m\in[1, M]$.
 \\ \textit{Initialization} : Randomly initialize the parameters of $M$ subnetworks $\theta_{m}, m\in[1, M]$ and the parameters of the shared MLP classifier $\theta_{P}$.
  \FOR {$j = 1$ to $\mathcal{N}$}
  \FOR {$b = 1$ to $\left \lfloor \frac{n}{n_b}  \right \rfloor$}
  \STATE Construct a training mini-batch by randomly selecting $n_b$ samples from $S$.
  \STATE Extract the representation $v_{i}^{m}$ for each sample $x_{i}^{m}$ in the mini-batch by forward propagation, where $m\in[1, M]$, and $i\in[1, n_b]$.
  \STATE For each $v_{i}^{m}$, acquire the class prediction $y_{i}^{m}$ by:\\
  $y_{i}^{m}=MLP(v_{i}^{m})$
  \STATE  Calculate the mini-batch training loss $L$ by Eq. \ref{eq:loss_combined}.
  \STATE Update the parameters of the entire network, where each part is updated by:\\
  
   \textbf{a) }Parameters of linear classifier $P$ is updated by minimizing $J$ in Eq. \ref{eq:loss_combined} with:\\
   $\quad \quad \quad \quad  \theta_{P} = \theta_{P} - \tau\frac{\partial J }{\partial\theta_{P}}$\\
   
   \textbf{b) }Parameters of the sub-networks, $\theta_{m}$, by minimizing $J$ with descending their stochastic gradient:\\
   $\quad \quad \quad \quad  \theta_{m} = \theta_{m} - \tau\frac{\partial J }{\partial\theta_{m}},\quad  m\in[1, M]$ \\
  
    \textbf{c) }Center of each class is updated by Eq. \ref{eq:center}.

  \ENDFOR
  \ENDFOR
 \end{algorithmic} 
 \end{algorithm}

\subsection{Framework Architecture}

The proposed loss function can be applied to various cross-modal retrieval tasks. To verify the effectiveness of the proposed loss function, we designed an end-to-end framework for 3D cross-modal retrieval task to jointly train multiple modalities including image, mesh, and point cloud. The overview of the proposed framework for 3D cross-modal retrieval is shown in Fig.~\ref{fig:framework}. As shown in the figure, there are three networks: $F(\theta)$ for image feature extraction, $G(\beta)$ for point cloud feature extraction, and $H(\gamma)$ for mesh feature extraction. Our framework can be easily extended to cases with more modalities or to different cross-modal retrieval tasks. We also verify the effectiveness of the proposed loss function on the image-text retrieval task. 

\textbf{3D cross-modal retrieval.} For 2D image feature extraction, we utilize ResNet18 \cite{he2016deep} as the backbone network with four convolution blocks, all with $3 \times 3$ kernels, where the number of kernels are $64$, $128$, $256$, and $512$, respectively. Unless specifically mentioned, after the global average pooling, a 512-dimensional final feature vector is acquired in all experiments. Dynamic graph convolutional neural network (DGCNN)\cite{dgcnn} is employed as the backbone model to capture point cloud features. DGCNN contains four EdgeConv blocks with the number of kernels set to $64$, $64$, $64$, and $128$. After the fourth EdgeConv block, a fully connected layer with $512$ neurons is used to extract point-specific features for each point and then a max-pooling layer is applied to extract global features for each object. MeshNet~\cite{feng2018meshnet} consists of 2 mesh convolution blocks, which achieved the state-of-the-art results for mesh retrieval, and is selected as the backbone to extract the features from mesh data. Two fully connected layers with size of $256$ and $40$ are employed to make classification predictions based on the 512-dimensional global features for all three modalities. The entire framework is trained from scratch for 3D cross-modal retrieval task with the proposed loss function. 


\section{Experiments}

\textbf{Datasets:} In this paper, ModelNet40 \cite{wu20153d} dataset is used for 3D cross-modal retrieval task. The ModelNet40 dataset is a 3D object benchmark and contains $12,311$ CAD models which belong to 40 different categories with $9,843$ used for training and $2,468$ for testing. Three modalities image, point cloud, and mesh are provided in this dataset.

\subsection{Experimental Setup:} 


\textbf{3D Cross-modal retrieval: } The proposed framework is jointly trained for all modalities including 2D image, point cloud and mesh on ModelNet40 \cite{wu20153d} dataset. We use SGD optimizer with an initial learning rate of $0.001$, the momentum of $0.9$, and weight decay of $0.001$. The network is updated with a mini-batch size of $96$ for $80,000$ iterations and the learning rate is decreased by a factor of $0.1$ every $20,000$ iterations. Data augmentation methods used for point cloud are random rotation of the point cloud between [$0$, $2\pi$] degrees along the up-axis as well as random jittering of the positions of every point by a Gaussian noise with zero mean and $0.02$ standard deviation. For images, the data is augmented using random cropping and random flipping with $0.5$ probability and for mesh, random rotation with a degree between [$0$, $2\pi$] is used.


\textbf{Evaluation Metrics:}  The evaluation results for all experiments are presented in terms of Mean Average Precision (mAP) score which is a classical performance evaluation criterion for cross-modal retrieval task \cite{zhuang2013supervised, feng2014cross, wang2017adversarial}. The mAP for retrieval task is defined to measure whether the retrieved data belong to the same class as the query (relevant) or do not (irrelevant). Given a query and a set of $R$ corresponding retrieved data ($R$ top-ranked data), the Average Precision is defined as

\begin{equation}
    \frac{1}{N} \sum_{r=1}^{R} p(r) \cdot \delta(r), 
\end{equation}
\noindent
where $N$ is the number of relevant data in the retrieved set, $p(r)$ is the precision of the first $r$ retrieved data, and $\delta(r)$ is the relevance of the $r$-th retrieved data (equal to 1 if relevant and 0 otherwise).

\subsection{3D Cross-modal Retrieval Task}

To evaluate the effectiveness of the proposed end-to-end framework, we conduct experiments on ModelNet40 dataset with three different modalities including  multi-view images, point cloud, and mesh. To thoroughly examine the quality of learned features, we conduct two types of retrieval tasks including in-domain retrieval when the source and target objects are from the same domain and cross-domain retrieval when they are from two different domains. When the target or source is from image domain, we evaluate the performance of multi-view images where the number of views is set to $1$, $2$ and $4$. The performance of our method for 3D in-domain and cross-modal retrieval tasks is shown in Table~\ref{tab:3D-performance}.

\begin{table}
	\caption{Performance of 3D in-domain and cross-modal retrieval task on ModelNet40 dataset in terms of mAP. When the target or source are from image domain, the results are reported for multi-view images: 1 view, 2 views, and 4 views denoted by v1, v2, and v4.}
	\begin{center}
	    \resizebox{0.48\textwidth}{!}{
    	\begin{tabular}{c|c|c c c}
    		\hline
            Source & Target & mAP-v1 & mAP-v2  & mAP-v4   \\
    		\hline
    		Image & Image  & 86.00  & 89.14 & 90.23  \\
    		Image & Mesh   & 87.31 & 88.98 & 89.59  \\
    		Image & Point Cloud  & 86.79 & 88.35 & 89.04 \\
    		Mesh & Image   & 85.96 & 87.83 & 88.11 \\
    		Point Cloud & Image  & 85.18 & 86.91 & 87.11  \\
    		Mesh & Mesh    & 88.59 & --- & ---  \\
    		Mesh & Point Cloud   & 87.37 & --- & --- \\
    		Point Cloud & Mesh   & 87.58 & --- & --- \\
    		Point Cloud & Point Cloud  & 87.04 & --- & --- \\
    		\hline
    	\end{tabular}}
	\end{center}
	\label{tab:3D-performance}
\end{table}

As shown in Table~\ref{tab:3D-performance}, the proposed framework achieves more than $85\%$ mAP for both in-domain and cross-domain retrieval tasks on ModelNet40 dataset. When the query or target are from the image-domain, the retrieval performance are significantly improved if more image views are used. Even though the cross-modal center loss is designed explicitly for learning modal invariant features, it can discriminate the features of different classes within the same domain and achieve more than $86\%$ mAP for Image2Image, Point2Point, and Mesh2Mesh in-domain retrieval tasks.

\subsection{Impact of Loss Function }

The three component of our proposed loss function are denoted as following: cross-entropy loss for each modality in the label space as $L_1$, cross-modal center loss in the universal representation space as $L_2$, and mean-square loss between features of different modalities as $L_3$. To further investigate the impact of each term, we evaluated different combinations for loss function including: 1) optimization with $L_1$, 2) jointly optimization with $L_1$ and $L_3$, 3) jointly optimization with $L_1$ and $L_2$ and 4) jointly optimization with $L_1$, $L_2$, and $L_3$. The four models are trained with the same setting and hyper-parameters, where the performance is shown in Table~\ref{tab:LossAblation}.

\begin{table}[ht]
	\caption{The ablation studies for loss function. $L_1$ is cross entropy loss, $L_2$ is cross-modal center loss, and $L_3$ is mean squared error loss. The number of views for images is fixed to $1$.}
	\begin{center}
	    \resizebox{0.48\textwidth}{!}{
		\begin{tabular}{l|c|c|c|c}
			\hline
			Loss   & $L_1$  & $L_1 + L_3$ &  $L_1 + L_2$ &  $L_1 + L_2 + L_3$  \\
			\hline
                Image2Image & 75.09 & 74.21 & 84.87 &  86.0  \\
                Image2Mesh & 75.38 & 75.86 & 86.7  & 87.31 \\
                Image2Point & 69.76 & 70.52 & 86.11 & 86.79 \\
                Mesh2Mesh & 75.53 & 76.36 & 88.83  & 88.59\\
                Mesh2Image & 75.2 & 74.76 & 85.66  & 85.96 \\
                Mesh2Point & 69.64 & 70.34 & 87.58  & 87.37\\
                Point2Point & 66.63 &68.18 & 86.89 & 87.04 \\
                Point2Image & 69.54 & 70.34& 84.76 & 85.18  \\
                Point2Mesh & 69.23 & 71.88& 87.69  & 87.58  \\
			\hline
		\end{tabular}}
	\end{center}
	\label{tab:LossAblation}
\end{table}


As illustrated in table ~\ref{tab:LossAblation}, we have
the following observations:

\begin{itemize}
\item The combination of $L_1$, $L_2$ and $L_3$ achieves the best performance for all cross-modal and in-domain retrieval tasks. 

\item As a baseline, the cross-entropy loss alone achieves relatively high mAP due to the shared head of the three modalities forcing the network to learn similar representations in the common space for different modalities of the same class. 

\item By combining the cross-modal center loss with cross-entropy loss, a consistent and significant improvement in mAP, between $7\%$ to $20\%$, could be achieved for different retrieval tasks, proving that the proposed cross-modal center loss could significantly reduce the cross-modal discrepancy.

\item Notably, the performance of Point2Mesh, Point2Point, and Mesh2Point retrieval tasks are improved by nearly $20\%$, which further validates the effectiveness of the proposed cross-modal center loss. 

\item Adding the MSE loss to cross-entropy and cross-modal center loss also slightly improves the performance (nearly $1\%$).
\end{itemize}


\subsection{Impact of Batch Size} 

The core idea of the proposed cross-modal center loss is to learn a unique center for each class and to minimize the distance of data from different modalities to the class center. However, the computation based on the whole dataset in each update is inefficient, even impractical \cite{wen2016discriminative}. Instead, the class centers are calculated by averaging the features in each mini-batch during the training. Therefore, the reliability of the features for each class is highly correlated with the batch size. Using large enough batch size provides sufficient samples for each class to find a reliable center while having a small batch size leads to unreliable centers. To analyze the impact of batch size on the performance, we conduct experiments on the ModelNet40 dataset with different batch sizes ($12, 24, 48, 96$). The results are shown in Table~\ref{tab:batchsize}. All models are trained with the same number of epochs and the same hyper-parameters.

\begin{table}[ht]
	\caption{The ablation studies for the batch size on the ModelNet40 dataset. The number of views for images is fixed to 4. Same number of epochs is used for all the experiments.}
	\begin{center}
		\begin{tabular}{l|c|c|c|c}
			\hline
			Loss  &12 &24  & 48 &96 \\
			\hline
			Image2Image   & 45.67 & 63.56   & 85.64 & 90.23 \\
			Image2Mesh    & 13.89 & 73.22   & 86.94 & 89.59 \\
			Image2Point   & 32.32 & 72.08   & 85.59 & 89.04 \\
			Mesh2Mesh     & 25.5 & 88.44    & 88.91 & 88.51 \\
			Mesh2Image    & 6.98 & 68.81   & 86.5 & 88.11 \\
			Mesh2Point    & 8.29 & 84.6   & 86.67 & 87.33 \\
			Point2Point   & 59.5 & 82.44   & 85.44 & 86.76 \\
			Point2Image   & 27.68 & 67.46  & 84.67 & 87.11 \\
			Point2Mesh    & 15.87 & 83.56    & 86.62 & 87.29 \\
			\hline
		\end{tabular}
	\end{center}
	\label{tab:batchsize}
\end{table}

\begin{figure}[ht]
\centering
\includegraphics[width=0.5\textwidth]{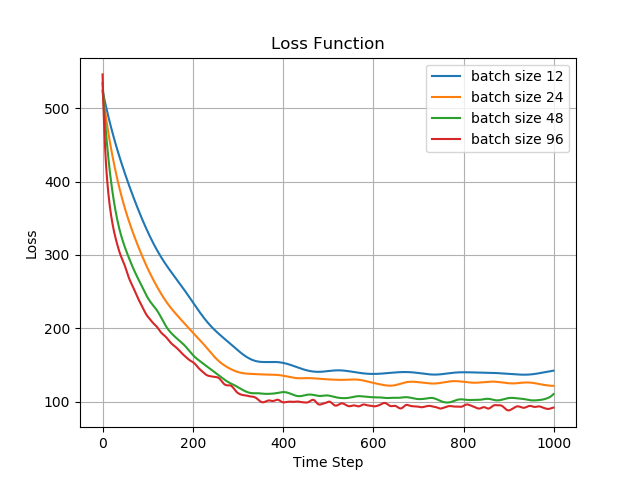}
\caption{The effect of batch size on cross-modal center loss. The curves are smoothed with Kernel Regression \cite{li2004cross}. Larger batch size significantly reduces the loss.}
\label{fig:loss-batch}
\end{figure}

As shown in Table~\ref{tab:batchsize}, changing the batch size from $12$ to $96$ significantly improves the mAP for all modalities. We also displayed the convergence of loss function with different batch sizes in Fig.~\ref{fig:loss-batch} for comparison. The figure shows that the large batch size leads to lower loss, suggesting that a larger batch size should be used for the proposed cross-modal center loss when possible.

\subsection{Comparison with Existing Methods on 3D Retrieval} \label{Comparison}

In this section, we compare the performance of our method with state-of-the-art supervised and self-supervised learning methods on 3D in-domain and cross-modal retrieval tasks. Since there is no supervised learning work on the 3D cross-modal retrieval task yet, we re-produce the current state-of-the-art method (DSCMR) designed for the image-text retrieval task and compare with its performance on ModelNet40 dataset. Since DSCMR was originally designed only for image-text retrieval, we extend it to three types of modalities and jointly trained it on the ModelNet40 dataset. The DSCMR employs MSE loss in the label space to learn discriminative features, and we found that the performance is very low on the ModelNet40 dataset. We replaced the MSE loss in the label space with the cross entropy loss and obtained a much stronger baseline (DSCMR+) on the ModelNet40 dataset. The performance is shown in Table~\ref{tab:Compare}. 

\begin{table}
	\caption{Comparison with state-of-the-art self-supervised and supervised methods on the ModelNet40 Dataset. The proposed method with joint training significantly outperforms all other methods on all retrieval tasks on the ModelNet40 dataset.}
	\begin{center}
	    \small
	    \resizebox{0.48\textwidth}{!}{
    	\begin{tabular}{l|c|c |c |c}
    		\hline
    		Retrieval$-$Views  & MVI \cite{MVI}  & DSCMR \cite{zhen2019deep}  & DSCMR+ &  Ours \\
    		\hline
    		Image2Image$-1$ & 57.9   &  56.22 &  73.35   & \textbf{86.00}    \\
    		Image2Image$-2$ & 60.55  &  60.36 & 75.87  & \textbf{89.14}  \\
    		Image2Image$-4$ & 61.92  &  62.71  & 77.82  & \textbf{90.23} \\
    		Image2Mesh$-1$ & 59.81  &  58.79  & 72.05  & \textbf{87.31}   \\
    		Image2Mesh$-2$ & 61.1   &  60.09  & 73.16  & \textbf{88.98}  \\
    		Image2Mesh$-4$ & 61.7   &  61.11  & 73.97  & \textbf{89.59} \\
    		Image2Point$-1$ & 59.56  &  50.64 & 68.81 & \textbf{86.79} \\
            Image2Point$-2$ & 60.76 & 51.35 & 69.58  & \textbf{88.35}\\ 
            Image2Point$-4$ & 61.38 & 52.07 & 70.35 & \textbf{89.04}\\
    		Mesh2Mesh$-\ast$ & 62.35 &  64.31 & 71.2 & \textbf{88.51}  \\
    		Mesh2Image$-1$&  59.06  &  52.22 &  71.12 & \textbf{85.96} \\
            Mesh2Image$-2$ & 60.73 & 55.62 & 72.63  & \textbf{87.83} \\
            Mesh2Image$-4$ & 61.66 & 57.09 & 73.75 & \textbf{88.11}\\
    		Mesh2Point$-\ast$ & 61.6 & 50.11 & 67.05  & \textbf{87.37} \\
    		Point2Point$-\ast$ & 62.12 & 56.55&  66.38 &  \textbf{87.04}  \\
    		Point2Image$-1$ &  59.05 &  53.77 &  68.59 & \textbf{85.18} \\
    		Point2Image$-2$ & 60.69  & 56.39 & 70.04  & \textbf{86.91} \\
    		Point2Image$-4$ & 61.56  & 58.03  & 71.23 & \textbf{87.11} \\
    		Point2Mesh$-\ast$ &  61.9   & 60.47 &  67.92  &  \textbf{87.58}  \\
    		\hline
    	\end{tabular}}
	\end{center}
	\label{tab:Compare}
\end{table}

We compare our results with the state-of-the-art self-supervised learning method, MVI \cite{MVI}, which learns the modal/view-invariant features by using cross-modal and cross-view invariant constraints. Our proposed model outperforms MVI by more than $20$\% in all in-domain and cross-modal retrieval tasks on the ModelNet40 dataset. When extending the state-of-the-art supervised method DSCMR on the ModelNet40 dataset with three modalities, the performance of DSCMR is much lower than our proposed method, and it is even slightly worse than the self-supervised learning method, MVI\cite{MVI}. However, by merely replacing the MSE loss in DSCMR with cross-entropy loss, the DSCMR+ could significantly improve the performance by more than $10\%$ while it is still much lower than the proposed method on all the retrieval tasks. These results further prove the effectiveness of the proposed method for cross-modal retrieval tasks.

\subsection{Qualitative Visualization}

\begin{figure*}[tb]
\centering
\includegraphics[width=\textwidth]{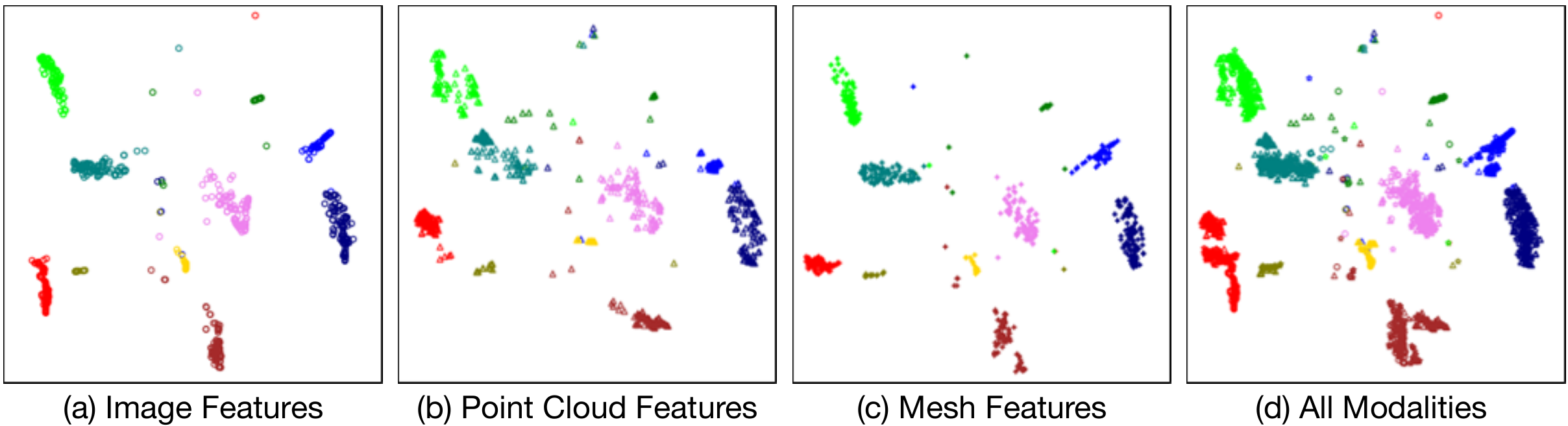}
\caption{The visualization results on the test data of the ModelNet40 dataset by using the t-SNE method \cite{maaten2008visualizing}. Each point in the figure represents one object, and objects from the same categories are rendered with the same color.}
\label{fig:TSNE}
\end{figure*}

\textbf{t-SNE Feature Embedding Visualization: } To investigate the effectiveness of the proposed model in the feature space, we employ the t-SNE method \cite{maaten2008visualizing} to embed the learned features of different modalities into a 2D plane. The visualization of features for the testing split of the ModelNet40 dataset with our method is shown in Fig.~\ref{fig:TSNE}. Fig.~\ref{fig:TSNE} (a), (b), and (c) show the features from the image, point cloud, and mesh modalities extracted by our proposed model, respectively. Fig.~\ref{fig:TSNE} (d) shows the mixed features of image, point cloud, and mesh modalities extracted by our model.

Fig.~\ref{fig:TSNE} (a), (b), and (c) show that the features are distributed as separated clusters, demonstrating that the proposed loss is able to discriminate the samples from different classes for each modality. From Fig.~\ref{fig:TSNE} (d), the features from three different modalities are mixed, showing that the features learned by the proposed framework in the universal space are indeed model-invariant.

\begin{figure*}[!tb]
\centering
\includegraphics[width = \textwidth]{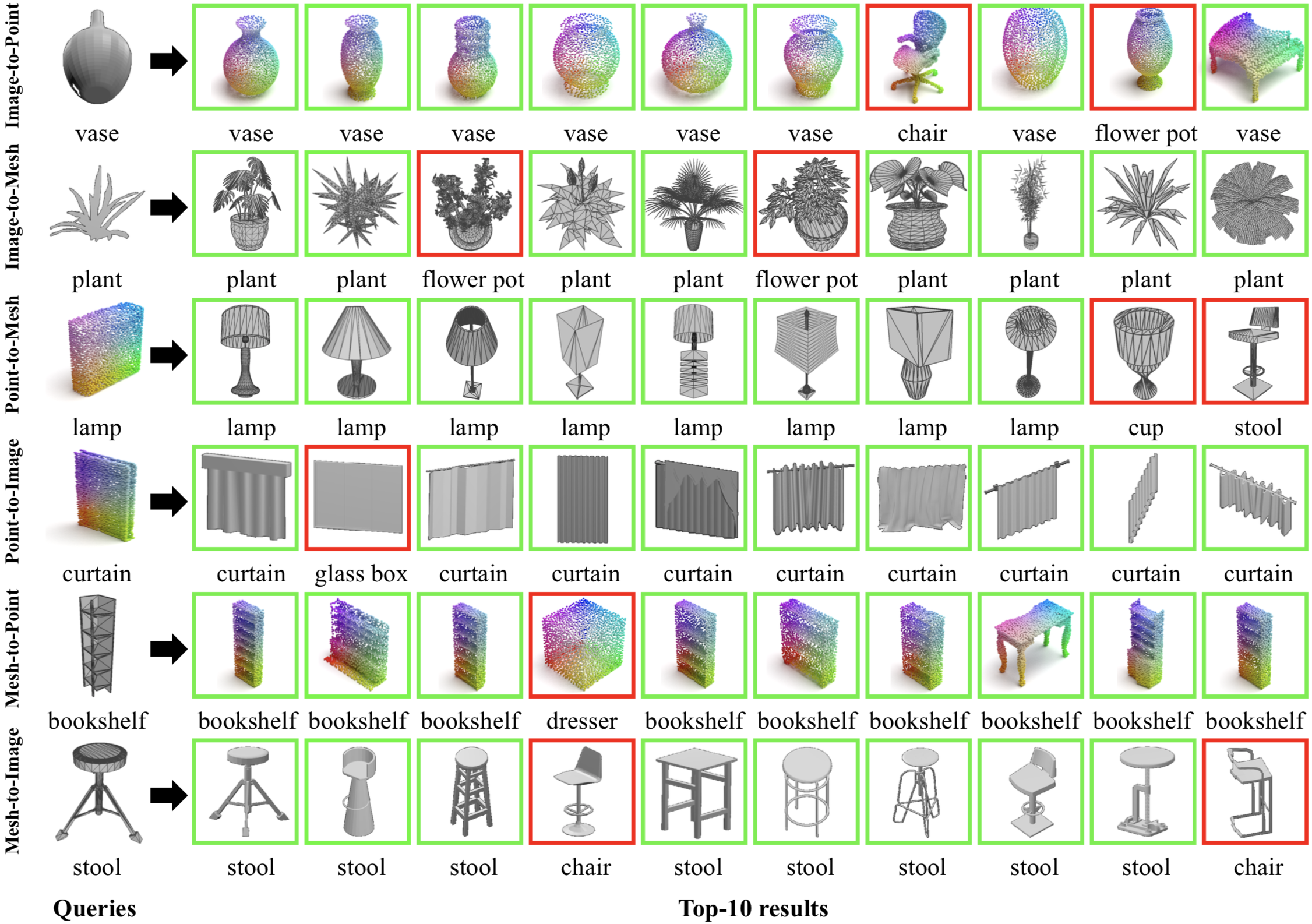}
\caption{Top-10 retrieval results of our model for six query samples on the ModelNet40 dataset. The green bounding boxes indicate that the images belong to the same category as the query, whereas the red bounding boxes indicate wrong matches.
}
\label{fig:3DResults}
\end{figure*}

\textbf{Cross-modal retrieval Visualization:} Fig.~\ref{fig:3DResults} shows the cross-modal retrieval samples for six different queries from ModelNet40 dataset. For each query, the euclidean distance over the normalized features is used to measure the similarity of data from different modalities. The Top-10 closest samples for each query are selected and visualized. The figure shows that objects with similar appearances are closer in the features space even though they are from different modalities, proving that the network indeed learned model-invariant features.

\section{Conclusion}

In this paper, we proposed a cross-modal center loss to learn discriminative and modal-invariant features for cross-modal retrieval tasks. The proposed cross-modal center loss significantly reduces the cross-modal discrepancy by minimizing the distances of features belonging to the same class across all modalities and can be used in conjunction with other loss functions. Extensive experiments have been conducted on retrieval tasks across multi-modalities, including image, 3D point cloud, and mesh data. The proposed framework significantly outperforms the state-of-the-art methods on the ModelNet40 dataset, validating the effectiveness of the proposed cross-modal center loss and the end-to-end framework.


{\small
\bibliographystyle{ieee}
\bibliography{root}

\begin{thebibliography}{10}\itemsep=-1pt

\bibitem{andrew2013deep}
G.~Andrew, R.~Arora, J.~Bilmes, and K.~Livescu.
\newblock Deep canonical correlation analysis.
\newblock In {\em International conference on machine learning}, pages
  1247--1255, 2013.

\bibitem{chang2015shapenet}
A.~X. Chang, T.~Funkhouser, L.~Guibas, P.~Hanrahan, Q.~Huang, Z.~Li,
  S.~Savarese, M.~Savva, S.~Song, H.~Su, et~al.
\newblock Shapenet: An information-rich 3d model repository.
\newblock {\em arXiv preprint arXiv:1512.03012}, 2015.

\bibitem{feng2014cross}
F.~Feng, X.~Wang, and R.~Li.
\newblock Cross-modal retrieval with correspondence autoencoder.
\newblock In {\em Proceedings of the 22nd ACM international conference on
  Multimedia}, pages 7--16, 2014.

\bibitem{feng2018meshnet}
Y.~Feng, Y.~Feng, H.~You, X.~Zhao, and Y.~Gao.
\newblock Meshnet: Mesh neural network for 3d shape representation.
\newblock {\em AAAI 2019}, 2018.

\bibitem{hadsell2006dimensionality}
R.~Hadsell, S.~Chopra, and Y.~LeCun.
\newblock Dimensionality reduction by learning an invariant mapping.
\newblock In {\em 2006 IEEE Computer Society Conference on Computer Vision and
  Pattern Recognition (CVPR'06)}, volume~2, pages 1735--1742. IEEE, 2006.

\bibitem{hanocka2019meshcnn}
R.~Hanocka, A.~Hertz, N.~Fish, R.~Giryes, S.~Fleishman, and D.~Cohen-Or.
\newblock Meshcnn: a network with an edge.
\newblock {\em ACM Transactions on Graphics (TOG)}, 38(4):1--12, 2019.

\bibitem{he2016deep}
K.~He, X.~Zhang, S.~Ren, and J.~Sun.
\newblock Deep residual learning for image recognition.
\newblock In {\em Proceedings of the IEEE conference on computer vision and
  pattern recognition}, pages 770--778, 2016.

\bibitem{hotelling1992relations}
H.~Hotelling.
\newblock Relations between two sets of variates.
\newblock In {\em Breakthroughs in statistics}, pages 162--190. Springer, 1992.

\bibitem{XMV}
L.~Jing, Y.~Chen, L.~Zhang, M.~He, and Y.~Tian.
\newblock Self-supervised feature learning by cross-modality and cross-view
  correspondences.
\newblock {\em arXiv preprint arXiv:2004.05749}, 2020.

\bibitem{MVI}
L.~Jing, Y.~Chen, L.~Zhang, M.~He, and Y.~Tian.
\newblock Self-supervised modal and view invariant feature learning.
\newblock {\em arXiv preprint arXiv:2005.14169}, 2020.

\bibitem{SelfSurvey}
L.~Jing and Y.~Tian.
\newblock Self-supervised visual feature learning with deep neural networks: A
  survey.
\newblock {\em IEEE Transactions on Pattern Analysis and Machine Intelligence},
  2020.

\bibitem{kan2015multi}
M.~Kan, S.~Shan, H.~Zhang, S.~Lao, and X.~Chen.
\newblock Multi-view discriminant analysis.
\newblock {\em IEEE transactions on pattern analysis and machine intelligence},
  38(1):188--194, 2015.

\bibitem{khosla2020supervised}
P.~Khosla, P.~Teterwak, C.~Wang, A.~Sarna, Y.~Tian, P.~Isola, A.~Maschinot,
  C.~Liu, and D.~Krishnan.
\newblock Supervised contrastive learning.
\newblock {\em arXiv preprint arXiv:2004.11362}, 2020.

\bibitem{volum4}
R.~Klokov and V.~Lempitsky.
\newblock Escape from cells: Deep kd-networks for the recognition of 3d point
  cloud models.
\newblock In {\em ICCV}, pages 863--872, 2017.

\bibitem{lei2019spherical}
H.~Lei, N.~Akhtar, and A.~Mian.
\newblock Spherical kernel for efficient graph convolution on 3d point clouds.
\newblock {\em arXiv preprint arXiv:1909.09287}, 2019.

\bibitem{li2019deepgcns}
G.~Li, M.~Muller, A.~Thabet, and B.~Ghanem.
\newblock Deepgcns: Can gcns go as deep as cnns?
\newblock In {\em Proceedings of the IEEE International Conference on Computer
  Vision}, pages 9267--9276, 2019.

\bibitem{li2004cross}
Q.~Li and J.~Racine.
\newblock Cross-validated local linear nonparametric regression.
\newblock {\em Statistica Sinica}, pages 485--512, 2004.

\bibitem{maaten2008visualizing}
L.~v.~d. Maaten and G.~Hinton.
\newblock Visualizing data using t-sne.
\newblock {\em Journal of machine learning research}, 9(Nov):2579--2605, 2008.

\bibitem{volum1}
D.~Maturana and S.~Scherer.
\newblock Voxnet: A 3d convolutional neural network for real-time object
  recognition.
\newblock In {\em IROS}, pages 922--928. IEEE, 2015.

\bibitem{nagrani2018seeing}
A.~Nagrani, S.~Albanie, and A.~Zisserman.
\newblock Seeing voices and hearing faces: Cross-modal biometric matching.
\newblock In {\em Proceedings of the IEEE conference on computer vision and
  pattern recognition}, pages 8427--8436, 2018.

\bibitem{peng2016cross}
Y.~Peng, X.~Huang, and J.~Qi.
\newblock Cross-media shared representation by hierarchical learning with
  multiple deep networks.

\bibitem{peng2017ccl}
Y.~Peng, J.~Qi, X.~Huang, and Y.~Yuan.
\newblock Ccl: Cross-modal correlation learning with multigrained fusion by
  hierarchical network.
\newblock {\em IEEE Transactions on Multimedia}, 20(2):405--420, 2017.

\bibitem{qi2017pointnet}
C.~R. Qi, H.~Su, K.~Mo, and L.~J. Guibas.
\newblock Pointnet: Deep learning on point sets for 3d classification and
  segmentation.
\newblock In {\em Proceedings of the IEEE conference on computer vision and
  pattern recognition}, pages 652--660, 2017.

\bibitem{volum5}
C.~R. Qi, H.~Su, M.~Nie{\ss}ner, A.~Dai, M.~Yan, and L.~J. Guibas.
\newblock Volumetric and multi-view cnns for object classification on 3d data.
\newblock In {\em CVPR}, pages 5648--5656, 2016.

\bibitem{qi2017pointnet++}
C.~R. Qi, L.~Yi, H.~Su, and L.~J. Guibas.
\newblock Pointnet++: Deep hierarchical feature learning on point sets in a
  metric space.
\newblock In {\em Advances in neural information processing systems}, pages
  5099--5108, 2017.

\bibitem{rupnik2010multi}
J.~Rupnik and J.~Shawe-Taylor.
\newblock Multi-view canonical correlation analysis.
\newblock In {\em Conference on Data Mining and Data Warehouses (SiKDD 2010)},
  pages 1--4, 2010.

\bibitem{schroff2015facenet}
F.~Schroff, D.~Kalenichenko, and J.~Philbin.
\newblock Facenet: A unified embedding for face recognition and clustering.
\newblock In {\em Proceedings of the IEEE conference on computer vision and
  pattern recognition}, pages 815--823, 2015.

\bibitem{su2015multi}
H.~Su, S.~Maji, E.~Kalogerakis, and E.~Learned-Miller.
\newblock Multi-view convolutional neural networks for 3d shape recognition.
\newblock In {\em Proceedings of the IEEE international conference on computer
  vision}, pages 945--953, 2015.

\bibitem{su2018deeper}
J.-C. Su, M.~Gadelha, R.~Wang, and S.~Maji.
\newblock A deeper look at 3d shape classifiers.
\newblock In {\em Proceedings of the European Conference on Computer Vision
  (ECCV)}, pages 0--0, 2018.

\bibitem{volum3}
M.~Tatarchenko, A.~Dosovitskiy, and T.~Brox.
\newblock Octree generating networks: Efficient convolutional architectures for
  high-resolution 3d outputs.
\newblock In {\em ICCV}, pages 2088--2096, 2017.

\bibitem{thomas2019kpconv}
H.~Thomas, C.~R. Qi, J.-E. Deschaud, B.~Marcotegui, F.~Goulette, and L.~J.
  Guibas.
\newblock Kpconv: Flexible and deformable convolution for point clouds.
\newblock In {\em Proceedings of the IEEE International Conference on Computer
  Vision}, pages 6411--6420, 2019.

\bibitem{wang2017adversarial}
B.~Wang, Y.~Yang, X.~Xu, A.~Hanjalic, and H.~T. Shen.
\newblock Adversarial cross-modal retrieval.
\newblock In {\em Proceedings of the 25th ACM international conference on
  Multimedia}, pages 154--162, 2017.

\bibitem{wang2015deep}
W.~Wang, R.~Arora, K.~Livescu, and J.~Bilmes.
\newblock On deep multi-view representation learning.
\newblock In {\em International Conference on Machine Learning}, pages
  1083--1092, 2015.

\bibitem{wang2019dynamic}
Y.~Wang, Y.~Sun, Z.~Liu, S.~E. Sarma, M.~M. Bronstein, and J.~M. Solomon.
\newblock Dynamic graph cnn for learning on point clouds.
\newblock {\em ACM Transactions on Graphics (TOG)}, 38(5):1--12, 2019.

\bibitem{dgcnn}
Y.~Wang, Y.~Sun, Z.~Liu, S.~E. Sarma, M.~M. Bronstein, and J.~M. Solomon.
\newblock Dynamic graph cnn for learning on point clouds.
\newblock {\em ACM Transactions on Graphics (TOG)}, 2019.

\bibitem{wen2016discriminative}
Y.~Wen, K.~Zhang, Z.~Li, and Y.~Qiao.
\newblock A discriminative feature learning approach for deep face recognition.
\newblock In {\em European conference on computer vision}, pages 499--515.
  Springer, 2016.

\bibitem{wu2019pointconv}
W.~Wu, Z.~Qi, and L.~Fuxin.
\newblock Pointconv: Deep convolutional networks on 3d point clouds.
\newblock In {\em Proceedings of the IEEE Conference on Computer Vision and
  Pattern Recognition}, pages 9621--9630, 2019.

\bibitem{wu20153d}
Z.~Wu, S.~Song, A.~Khosla, F.~Yu, L.~Zhang, X.~Tang, and J.~Xiao.
\newblock 3d shapenets: A deep representation for volumetric shapes.
\newblock In {\em Proceedings of the IEEE conference on computer vision and
  pattern recognition}, pages 1912--1920, 2015.

\bibitem{zhai2013learning}
X.~Zhai, Y.~Peng, and J.~Xiao.
\newblock Learning cross-media joint representation with sparse and
  semisupervised regularization.
\newblock {\em IEEE Transactions on Circuits and Systems for Video Technology},
  24(6):965--978, 2013.

\bibitem{zhen2019deep}
L.~Zhen, P.~Hu, X.~Wang, and D.~Peng.
\newblock Deep supervised cross-modal retrieval.
\newblock In {\em Proceedings of the IEEE Conference on Computer Vision and
  Pattern Recognition}, pages 10394--10403, 2019.

\bibitem{zhou2018visual}
Y.~Zhou, Z.~Wang, C.~Fang, T.~Bui, and T.~L. Berg.
\newblock Visual to sound: Generating natural sound for videos in the wild.
\newblock In {\em Proceedings of the IEEE Conference on Computer Vision and
  Pattern Recognition}, pages 3550--3558, 2018.

\bibitem{zhuang2013supervised}
Y.~T. Zhuang, Y.~F. Wang, F.~Wu, Y.~Zhang, and W.~M. Lu.
\newblock Supervised coupled dictionary learning with group structures for
  multi-modal retrieval.
\newblock In {\em Twenty-Seventh AAAI Conference on Artificial Intelligence},
  2013.

\end{thebibliography}
}

\end{document}